\def\year{2019}\relax
\documentclass[letterpaper]{article} 
\usepackage{aaai19}  
\usepackage{times}  
\usepackage{helvet}  
\usepackage{courier}  
\usepackage{url}  
\usepackage{graphicx}  
\usepackage{microtype}
\usepackage{subfigure}
\usepackage[utf8]{inputenc} 
\usepackage[T1]{fontenc}    
\usepackage{hyperref}       
\usepackage{booktabs}       
\usepackage{amsfonts}       
\usepackage{nicefrac}       
\usepackage{amsmath}    
\usepackage{amssymb}    
\usepackage{multicol,lipsum}
\usepackage{multirow}
\usepackage{placeins}
\usepackage{tikz} 
\usetikzlibrary{arrows}
\usetikzlibrary{arrows.meta}
\usetikzlibrary{shapes}
\usetikzlibrary{decorations.pathreplacing}
\usepackage{xcolor}
\usepackage{color}
\usepackage{booktabs} 
\usepackage{lipsum}
\usepackage{mathtools}
\usepackage{cuted}
\usepackage{algorithm}
\usepackage{algorithmic}
\usepackage{todonotes}
\newtheorem{theorem}{Theorem}
\pdfoutput=1
\usepackage{pdfpages}

\newtheorem{definition}[theorem]{Definition}
\DeclareMathOperator*{\argmax}{arg\,max}

\DeclareMathOperator{\E}{\mathbb{E}}
\begin{document}
%
\title{Response Characterization for Auditing Cell Dynamics in Long Short-term Memory Networks}
\author{Ramin M. Hasani$^{1*}$, Alexander Amini$^{2*}$,Mathias Lechner$^{1~3}$, Felix Naser$^{2}$, Radu Grosu$^{1}$, Daniela Rus$^{2}$\\
$^{1}$Cyber Physical Systems (CPS), Technische Universit\"{a}t Wien (TU Wien), 1040 Vienna, Austria\\
$^{2}$ Computer Science and Artificial Intelligence Lab (CSAIL), Massachusetts Institute of Technology (MIT), Cambridge, USA\\
$^{3}$Institute of Science and Technology (IST), 3400 Klosterneuburg, Austria\\
$^{*}$Equal Contributions
}
\maketitle
\begin{abstract}
In this paper, we introduce a novel method to interpret recurrent neural networks (RNNs), particularly long short-term memory networks (LSTMs) at the cellular level. 
We propose a systematic pipeline for interpreting individual hidden state dynamics within the network using response characterization methods. 
The ranked contribution of individual cells to the network's output is computed by analyzing a set of interpretable metrics of their decoupled step and sinusoidal responses. 
As a result, our method is able to uniquely identify neurons with insightful dynamics, quantify relationships between dynamical properties and test accuracy through ablation analysis, and interpret the impact of network capacity on a network's dynamical distribution. 
Finally, we demonstrate generalizability and scalability of our method by evaluating a series of different benchmark sequential datasets.
\end{abstract}

\section{Introduction}
\label{sec:introduction}

A key challenge for modern deep learning architectures is that of robust interpretation of its hidden dynamics and how they contribute to the system's decision making ability as a whole. 
%
Many safety critical applications of deep neural networks (NNs), such as robotic control and autonomous driving \cite{mnih2015human,levine2016end,bojarski2016end}, require metrics of explainability before they are deployed into the real world. 
In particular, interpreting the dynamics of recurrent neural networks (RNNs), which can process sequential data, and are vastly used in such safety critical domains requires careful engineering of network architecture \cite{karpathy2015visualizing}. This is because investigating their behavior enables us to reason about their hidden state-dynamics in action and thus design better models.

The hidden state representations of long short-term memory (LSTM) networks \cite{hochreiter1997long}, a subset of RNNs with explicit gating mechanisms, have been evaluated by gate-ablation analysis \cite{chung2014empirical,greff2017lstm} and feature visualization in linguistics \cite{karpathy2015visualizing,strobelt2018lstmvis}. 
While these studies provide criteria for networks with interpretable cells, they are limited to feature visualization techniques, focus on hidden state dynamics in networks for text analysis, and thus suffer from poor generalizability. A robust, systematic method for assessing RNN dynamics across all sequential data modalities has yet to be developed.%

In this paper, we introduce a novel methodology to predict and interpret the hidden dynamics of LSTMs at the individual cell and global network level. 
We utilize response characterization techniques \cite{oppen1983}, wherein a dynamical system is exposed to a controlled set of input signals and the associated outputs are systematically characterized. 
Concretely, we present a systematic testbench to interpret the relative contributions, response speed, and even the phase shifted nature of learned LSTM models. 
%
To analyze hidden state dynamics, we isolate individual LSTM cells from trained networks and expose them to defined input signals such as step and sinusoid functions. Through evaluation of output attributes, such as response settling time, phase-shift, and amplitude, we demonstrate that it is possible to predict sub-regions of the network dynamics, rank cells based on their relative contribution to network output, and thus produce reproducible metrics of network interpretability.

For example, step response settling time delineates cells with fast and slow response dynamics. In addition, by considering the steady-state value of the cellular step response and the amplitude of the sinusoid response, we are able to identify cells that significantly contribute to a network's decision. We evaluate our methodology on a range of sequential datasets and demonstrate that our algorithms scale to large LSTM networks with millions of parameters.


The key contributions of this paper can be summarized as follows:
\begin{enumerate}
\item Design and implementation of a novel and lightweight algorithm for systematic LSTM interpretation based on response characterization; 
\item Evaluation of our interpretation method on four sequential datasets including classification and regression tasks; and
\item Detailed interpretation of our trained LSTMs on the single cell scale via distribution and ablation analysis as well as on the network scale via network capacity analysis. 
\end{enumerate}

First, we discuss related work in Sec. \ref{sec:related-work} and then introduce the notion of RNNs as dynamic systems in Sec.~\ref{sec:rnn-dynamic-systems}. Sec. \ref{sec:response-characterization} presents our algorithm for response characterization and defines the extracted interpretable definitions. Finally, we discuss the interpretations enabled by this analysis in Sec. \ref{sec:results}  through an series of experiments, and provide final conclusions of this paper in Sec. \ref{sec:conclusion}.



\section{Related Work}
\label{sec:related-work}

\textbf{Deep Neural Networks Interpretability -}
A number of impactful attempts have been proposed for interpretation of deep networks through feature visualization \cite{erhan2009visualizing,zeiler2014visualizing,yosinski2015understanding,karpathy2015visualizing,strobelt2018lstmvis,bilal2018convolutional}. 
Feature maps can be empirically interpreted at various scales using neural activation analysis \cite{olah2018building}, where the activations of hidden neurons or the hidden-state of these neurons is computed and visualized. Additional approaches try to understand feature maps by evaluating attributions \cite{simonyan2013deep,fong2017interpretable,kindermans2017patternnet,sundararajan2017axiomatic}. Feature attribution is commonly performed by computing saliency maps (a linear/non-linear heatmap that quantifies the contribution of every input feature to the final output decision). The contributions of hidden neurons, depending on the desired level of interpretability, can be highlighted at various scales ranging from individual cell level, to the channel and spatial filter space, or even to arbitrary groups of specific neurons \cite{olah2018building}. A dimensionality reduction method can also be used to abstract from high dimensional feature maps into a low dimensional latent space representation to qualitatively interpret the most important underlying features \cite{maaten2008visualizing,amini2018variational}. However, these methods often come with the cost of decreasing cell-level auditability. 

Richer infrastructures have been recently developed to reason about the network's intrinsic kinetics. LSTMVis \cite{strobelt2018lstmvis}, relates the hidden state dynamics patterns of the LSTM networks to similar patterns observed in larger networks to explain an individual cell's functionality. A systematic framework has also been introduced that combines interpretability methodologies across multiple network scales \cite{olah2018building}. This enables exploration over various levels of interpretability for deep NNs; however, there is still space to incorporate more techniques, such as robust statistics \cite{koh2017understanding}, information theory approaches \cite{shwartz2017opening}, and response characterization methods which we address in this paper.

\textbf{Recurrent Neural Networks Interpretability -}
Visualization of the hidden-state of a fixed-structure RNNs on text and linguistic datasets identifies interpretable cells which have learned to detect certain language syntaxes and semantics \cite{karpathy2015visualizing,strobelt2018lstmvis}. RNNs have also been shown to learn input-sensitive grammatical functions when their hidden activation patterns were visualized \cite{kadar2015lingusitic,kadar2017representation}. Moreover, gradient-based attribution evaluation methods were used to understand the RNN functionality in localizing key words in the text. While these techniques provide rich insight into the dynamics of learned linguistics networks, the interpretation of the network often requires detailed prior knowledge about the data content. Therefore, such methods may face difficulties in terms of generalization to other forms of sequential data such as time-series which we focus on in our study. 

Another way to build interpretability for RNNs is using the attention mechanism where the network architecture is constrained to attend to a particular parts of the input space. RNNs equipped with an attention mechanism have been successfully applied in image captioning, the fine-alignments in machine translation, and text extraction from documents \cite{hermann2015teaching}. Hidden-state visualization is a frequently shared property of all of these approaches in order to effectively understand the internals of the network. Hudson et al. \cite{hudson2018compositional} also introduced Memory, Attention, and Composition (MAC) cells which can be used to design interpretable machine reasoning engines in an end-to-end fashion. MAC is able to perform highly accurate reasoning, iteratively directly from the data. However, application of these modification to arbitrary network architectures is not always possible, and in the case of LSTM specifically, the extension is not possible in the current scope of MAC.

\textbf{Recurrent Neural Networks Dynamics-} Rigorous studies of the dynamical systems properties of RNNs, such as their activation function's independence property (IP) \cite{albertini1994uniqueness}, state distinguishability \cite{albertini1995recurrent}, and observability \cite{garzon1994observability,garzon1999dynamical} date back to more than two decades. Thorough analyses of how the long term dynamics are learned by the LSTM networks has been conducted in \cite{hochreiter1997long}. Gate ablation analysis on the LSTM networks has been performed to understand cell's dynamics \cite{greff2017lstm,chung2014empirical}. We introduce the response characterization method, as a novel building block  to understand and reason about LSTM hidden state dynamics. 

\section{Dynamics of Recurrent Neural Networks}
\label{sec:rnn-dynamic-systems}

In this section, we briefly we recap kinetics of RNNs. We denote the global dynamics of the hidden state values as $h_{t}^l$, with $t \in \{1..T\}$ denoting the time, and $l \in \{1..L\}$ representing the layers of the neural network. A \textit{vanilla recurrent neural network} (RNN) can be formulated as \cite{pascanu2013difficulty,karpathy2015visualizing}:
\begin{equation}
	h_{t}^l = \tanh \left( W^l  \left(\begin{array}{c} h_{t}^l \\ h_{t-1}^l \end{array}\right)\right),
\end{equation}
where $W^{l~[n\times2n]}$ shows the weight matrix. $h_{t}^0$ retains an input vector $x_t$ and $h_{t}^L$ holds a vector at the last hidden layer, $L$, that is mapped to an output vector $y_t$ which is ultimately the function of all input sequence $\{x_1,\dots,x_T\}$. 

RNNs are formulated as control dynamical systems in the form of the following differential equation (For the sake of notation simplicity, we omit the time argument, $t$): 

\begin{equation}
	\dot h = \sigma(Rh + Wx),~~~y = Cx,
\end{equation}

where $h$ denotes its internal state ($'~\dot{~}~'$ illustrates time-shift or time derivative for the discrete and continuous-time systems, respectively), $x$ stands for the input, and $R^{[n \times n]}$, $W^{[n \times m]}$ and $C^{[p \times n]}$ are real matrices representing recurrent weights, input weights and the output gains, respectively. $\sigma:\mathbb{R}\rightarrow\mathbb{R}$ indicates the activation function. In the continuous setting, $\sigma$ should be locally Lipschitz (see \cite{albertini1993neural} for a more detailed discussion). 

\subsection{Long Short-term Memory}

Long short term Memory (LSTM) \cite{hochreiter1997long}, are gated-recurrent neural networks architectures specifically designed to tackle the training challenges of RNNs. In addition to memorizing the state representation, they realize three gating mechanisms to read from input ($i$), write to output ($o$) and forget what the cell has stored ($f$). Activity of the cell can be formulated as follows \cite{greff2017lstm}: 

\begin{align}
&c^l_t = z \odot i + f \odot c^l_{t-1} \\
&y^l_t = o \odot \tanh(c^l_t)\\
&\begin{pmatrix}z\\i\\f\\o\end{pmatrix} =
\begin{pmatrix}\tanh\\\mathrm{\sigma}\\\mathrm{\sigma}\\\mathrm{\sigma}\end{pmatrix}
W^l \begin{pmatrix}y^{l - 1}_t\\y^l_{t-1}\end{pmatrix}
\end{align}

where $c^l_t$ is layer $l$'s cell state at time $t$, $W^{4n*2n}$ is the weight matrix, $z$ stands for the input block, and $y^l_t$ denotes the cell's output state. 

For analytical interpretation of a dynamical system, the first necessary condition is to check its observability property. A dynamical system is observable if there is some input sequence that gives rise to distinct outputs for two different initial states at which the system is started \cite{sontag2013mathematical}. Observable systems realize unique internal parameter settings \cite{albertini1994uniqueness}. One can then reason about that parameter setting to interpret the network for a particular input profile. Information flow in LSTM networks carries on by the composition of static and time-varying dynamical behavior. This interleaving of building blocks makes a complex partially-dependent sets of nonlinear dynamics that are hard to analytically formulate and to verify their observability properties \footnote{Read more about the observability matters of RNNs and LSTMs in the Supplementary Materials. Note that a more analytically rigorous investigation that focuses on the observability of LSTM networks, is required and will be the focus of our continued effort.}. As an alternative, in this paper we propose a technique for finding sub-regions of hidden observable dynamics within the network with a quantitative and systematic approach by using response characterization. 

\section{Methodology for Response Characterization of LSTM cells} 
\label{sec:response-characterization}

\begin{figure*}[t!]
  \centering
  \includegraphics[width=1\linewidth]{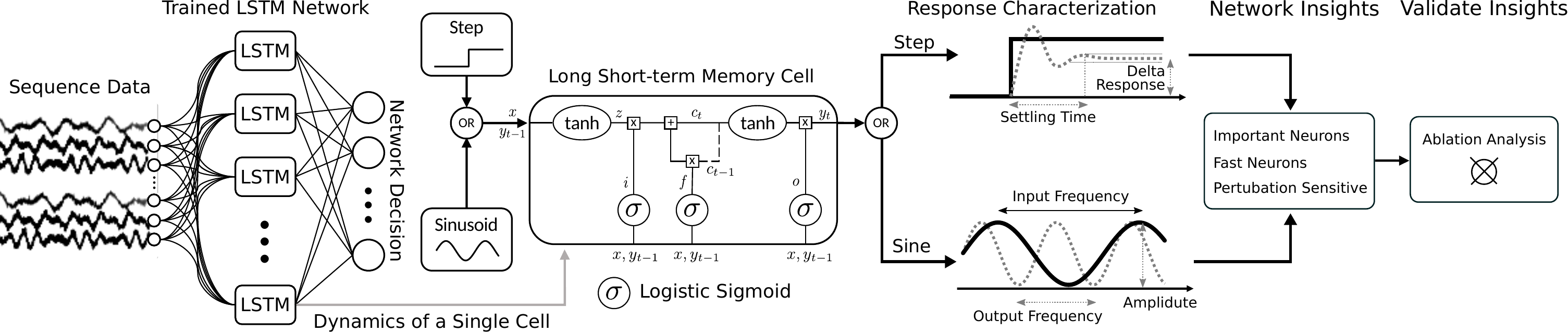}
  \caption{\textbf{Response characterization method for LSTM cells.} We take individual LSTM cells from a trained network, and characterize their step and sinusoidal response. These responses predict quantitative and interpretable measures for the dynamics of the single units within the network. We then validate the predictions by performing a neuronal ablation analysis.}
  \label{fig:schematic}
\end{figure*}

In this section, we explore how response characterization techniques can be utilized to perform systematic, quantitative, and interpretable understanding of LSTM networks on both a macro-network and micro-cell scale. By observing the output of the system when fed various baseline inputs, we enable a computational pipeline for reasoning about the dynamics of these hidden units. Figure~\ref{fig:schematic} provides a schematic for our response characterization pipeline. From a trained LSTM network, comprising of $M$ LSTM units, we isolate individual LSTM cells, and characterize their output responses based on a series of interpretable response metrics. We formalize the method as follows:

\begin{definition}
	Let G, be a trained LSTM network with $M$ hidden LSTM units. Given the dynamics of the training dataset (number of input/output channels, the main frequency components, the amplitude range of the inputs), we design specific step and sinusoidal inputs to the network, and get the following insights about the dynamics of the network at multi-scale resolutions:
	\begin{itemize}
		\item the relative strength or contribution of components within the network; 
		\item the reactiveness of components to sudden changes in input; and 
		\item the phase alignment of the hidden outputs with respect to the input.
	\end{itemize}
\end{definition}

Specifically, we analyze the responses of 
(1) the step input and 
(2) the sinusoidal input. We use the classic formulations for each of these input signals wherein (1) step: $x_t = \left[\left[t>\frac{T}{2}\right]\right]$; and (2) sinusoid: $x_t = \sin\left(2 \pi f\, t\right)$; 
where $[[\cdot]]$ represents the mathematical indicator function.


Across a network of LSTM units we can approximate sub-regions of the dynamics of a single cell, $u$, by extracting the input and recurrent weights corresponding to that individual cell. We then define a sub-system consisting of just that single cell and subsequently feed one of our baseline input signals, $x_t~\forall_{t\in\{1..T\}}$ to observe the corresponding output response, $y_t$. In the following, we define the interpretable response metrics for the given basis input used in this study: 

\begin{definition}
The \textbf{initial} and \textbf{final response} of the step response signal is the starting and steady state responses of the system respectively, while the \textbf{response output change} represents their relative difference.
\end{definition}

\textit{Response output change} or the delta response for short determines the strength of the LSTM unit with a particular parameter setting, in terms of output amplitude. This metric can presumably detect significant contributor units to the network's decision. 

\begin{definition}
The \textbf{settling time} of the step response is elapsed time from the instantaneous input change to when the output lies within a 90\% threshold window of its final response.
\end{definition}

Computing the \textit{settling time} for individual LSTM units enables us to discover ``fast units'' and ``slow units''. This leads to the prediction of active cells when responding to a particular input profile.

\begin{definition}
The \textbf{amplitude} and \textbf{frequency} of a cyclic response signal is the difference in output and rate at which the response output periodically cycles. The response frequency, $\hat f$, is computed by evaluating the highest energy component of the power spectral density: $\hat f=\argmax S_{yy}(f)$.
\end{definition}

The \textit{amplitude} metric enables us to rank LSTM cells in terms of significant contributions to the output. This criteria is specifically effective in case of trained RNNs on datasets with a cyclic nature. Given a sinusoidal input, phase-shifts and phase variations expressed at the unit's output, can be captured by evaluating the \textit{frequency} attribute. 

\begin{definition}
The \textbf{correlation} of the output response with respect to the input signal is the dot product between the unbiased signals: $\sum_{t=1}^T (x_t - \E[x]) \cdot (y_t-\E[y])$
\end{definition}

The \textit{correlation} metric correspondes to the phase-alignments between input and output of the LSTM unit.




Systematic computation of each of the above responses metrics for a given LSTM dynamics, enables reasoning on the internal kinetics of that system. Specifically, a given LSTM network can be decomposed into its individual cell components, thus creating many smaller dynamical systems, which can be analyzed according to their individual response characterization metrics. Repeating this process for each of the cells in the entire network creates two scales of dynamic interpretability. Firstly, on the individual cell level within the network to identify those which are inherently exhibiting \textit{fast} vs \textit{slow} responses to their input, quantify their relative contribution towards the system as a whole, and even interpret their underlying phase-shift and alignment properties. 
Secondly, in addition to characterizing responses on the cell level we also analyze the effect of network capacity on the dynamics of the network as a whole.
Interpreting hidden model dynamics is not only interesting as a deployment tool but also as a debugging tool to pinpoint possible sources of undesired dynamics within the network. 

While one can use these response characterization techniques to interpret individual cell dynamics, this analysis can also be done on the aggregate network scale. After computing our response metrics for all decoupled cells independently we then build full distributions over the set of all individual pieces of the network to gain understanding of the dynamics of the network as a whole. This study of the response metric distributions presents another rich representation for reasoning about the dynamics, no longer at a local cellular scale, but now, on the global network scale. 

\section{Experimental Results}
\label{sec:results}

In the following section, we provide concrete results of our system in practice to interpret the dynamics of trained LSTMs for various sequence modeling tasks. We present our computed metric response characteristics both on the decoupled cellular level as well as the network scale, and provide detailed and interpretable reasoning for these observed dynamics. We chose four benchmark sequential datasets and trained on various sized LSTM networks ranging from $32$ to $320$ LSTM cell networks. The results and analysis presented in this section demonstrate applicability of our algorithms to a wide range of temporal sequence problems and scalability towards deeper network structures.


We start by reasoning how our response characterization method can explain the hidden-state dynamics of learned LSTM networks for a sequential MNIST dataset and extend our findings to three additional datasets. We perform an ablation analysis and demonstrate how some of our metrics find cells with significant contributions to the network's decision, across all datasets. 

\subsection{Response characterization metrics predict insightful dynamics for individual cells}

\begin{figure*}[t!]
  \centering
  \includegraphics[width=\linewidth]{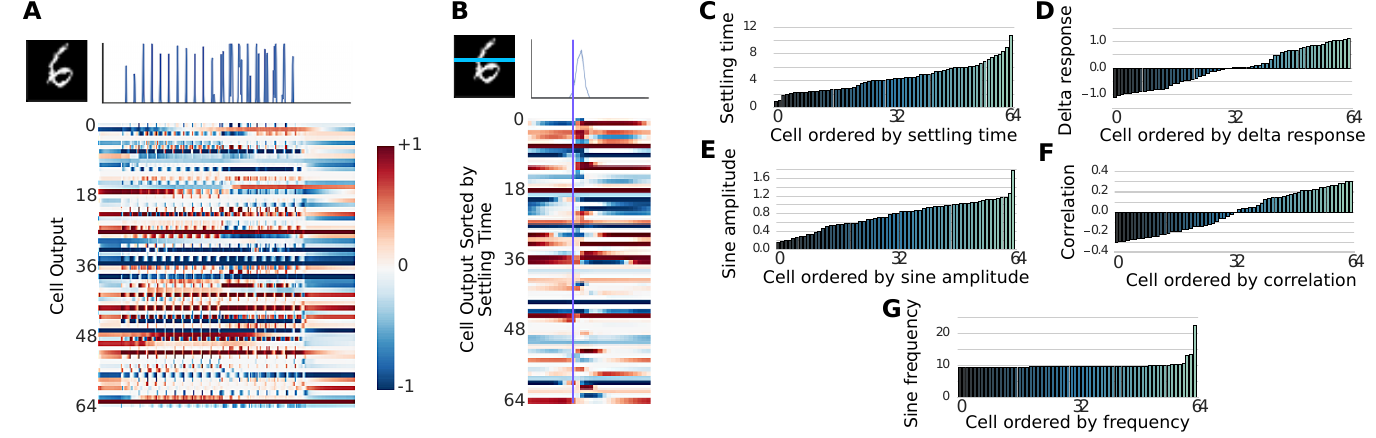}
  \caption{\textbf{Cell level interpretation of sequential MNIST.} A) An example sequence for digit 6 together with the network dynamics for a 64-neuron LSTM network. B) One slice sequence from digit 6 and its underlying network dynamics sorted for the settling time attribute. C) Settling time distribution. D) Delta response distribution. E) Sine-wave amplitude distribution. F) Correlation distribution G) Sine-frequency distribution.}
  \label{fig:response-char}
\end{figure*}

\begin{table*}[h!]
\scriptsize 
  \caption{\textbf{Hidden dynamic distributions by dataset.} Systematic interpretation of internal dynamics distributions (mean and variance) of $128$ cell LSTMs trained on various different benchmark datasets. The table shows the global speed and amplitude of the activity of network in terms of dynamical properties of the response characterization metrics. }
  \label{dataset-comparison}
  \centering
  \begin{tabular}{llllll}
      \toprule
      & \multicolumn{2}{c}{Step Response} & \multicolumn{3}{c}{Sinusoidal Response}                   \\
      \cmidrule{2-3}  \cmidrule{4-6}  
\textbf{Dataset} & \textbf{Settle Time} & \textbf{Output Change} & \textbf{Amplitude} & \textbf{Correlation} & \textbf{Frequency}  \\
    \midrule
Sequential-MNIST~\cite{lecun1998gradient}    & $6.96 \pm 4.08$  & $-0.04 \pm 0.58$  & $0.73 \pm 0.30$  & $0.17 \pm 0.08$  & $9.83 \pm 0.46$  \\
S\&P 500 Stock~\cite{stock}		             & $5.62 \pm 1.73$  & $0.02 \pm 0.16$   & $0.31 \pm 0.05$  & $0.03 \pm 0.02$  & $2.86 \pm 2.19$  \\
CO$_2$ Concentrations~\cite{co2}    	     & $5.65 \pm 1.64$  & $0.01 \pm 0.12$   & $0.27 \pm 0.04$  & $0.03 \pm 0.01$  & $9.83 \pm 0.08$  \\
Protein Sequencing~\cite{qian1988predicting} & $7.96 \pm 6.65$  & $0.08 \pm 0.54$   & $0.68 \pm 0.22$  & $2.07 \pm 1.21$  & $10.36 \pm 1.65$ \\
\bottomrule
\end{tabular}
\end{table*}

We start by training an LSTM network with $64$ hidden LSTM cells to classify a sequential MNIST dataset. Inputs to the cells are sequences of length $784$ generated by stacking the pixels of the $28\times28$ hand-writing digits, row-wise (cf. Fig. \ref{fig:response-char}A) and the output is the digit classification. 
Individual LSTM cells were then isolated and their step and sine-response were computed for the attributes defined formerly (cf. Fig. \ref{sec:response-characterization}). Fig. \ref{fig:response-char}C-G represent the distribution of cell activities, ranked by the specific metrics. The distribution of the settling time of the individual LSTM cells from a trained network, predicts low time-constant, (fast) cells (the first 20 neurons), and high-time constant (slow) cells (neurons 55-64) (Fig. \ref{fig:response-char}C). This interpretation allows us to indicate fast-activated/deactivated neurons at fast and slow phases of a particular input sequence. This is validated in Fig. \ref{fig:response-char}B, where the output state of individual LSTM cells are visually demonstrated when the network receives a sequence of the digit $6$. The figure is sorted in respect to the predicted settling time distribution. We observe that \textit{fast-cells} react to fast-input dynamics almost immediately while \textit{slow-cells} act in a slightly later phase. This effect becomes clear as you move down the heatmap in Fig. \ref{fig:response-char}B and observe the time difference from the original activation. 

The distribution of the delta-response, indicates inhibitory and excitatory dynamics expressed by a 50\% ratio (see Fig. \ref{fig:response-char}D). This is confirmed by the input-output correlation criteria, where almost half of the neurons express antagonistic behavior to their respective sine-wave input (Fig. \ref{fig:response-char}F). The sine-frequency distribution depicts that almost 90\% of the LSTM cells kept the phase, nearly aligned to their respective sine-input, which indicates existence of a linear transformation. A few cells learned to establish a faster frequencies than their inputs, thereby realizing phase-shifting dynamics (Fig. \ref{fig:response-char}G). The sine-amplitude distribution in Fig. \ref{fig:response-char}E demonstrates that the learned LSTM cells realized various amplitudes that are almost linearly increasing. The ones with a high amplitude can be interpreted as those maximally contributing to the network's decision. In the following sections, we investigate the generalization of these effects to other datasets.

\subsection{Generalization of response metrics are to other sequential datasets}

We trained LSTM networks with $128$ hidden cells, for four different temporal datasets: sequential MNIST \cite{lecun1998gradient}, S\&P 500 stock prices \cite{stock} and CO$_2$ concentration for the Mauna Laua volcano \cite{co2} forecasting, and classification of protein secondary structure \cite{qian1988predicting}. Learned networks for each dataset are denoted seq-MNIST, Stock-Net, CO$_2$-Net and Protein-Net. Table \ref{dataset-comparison} summarizes the statistics for all five metrics with the network size of $128$. It represents the average cell response metric attributes for various datasets and demonstrates the global speed and amplitude of the activity of network in terms of dynamical properties of the response characterization metrics. 

\begin{figure*}[t!]
  \centering
  \includegraphics[width=\linewidth]{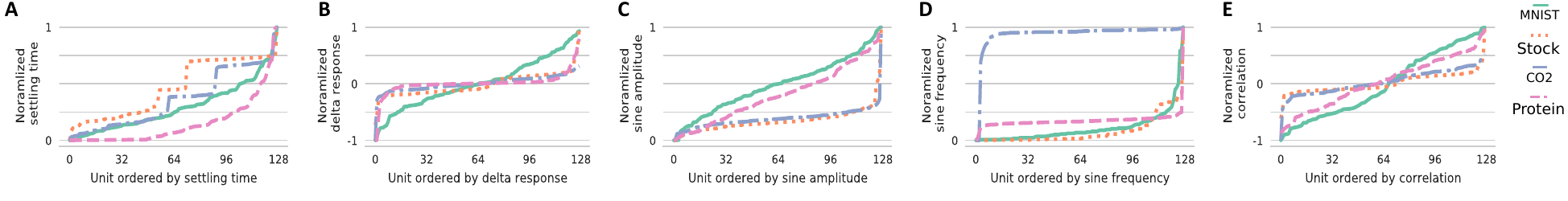}
  \caption{\textbf{Cell level response distributions.} (A-E) Response characterization metrics for networks with $128$ individually ranked LSTM cells. The analyses predict A) cells with fast-dynamics and slow dynamic, (B and C) cells that are significantly contributing to the network decision, D) cells that realize phase shifting dynamics, and E) cells that are excitatory or inhibitory.}
  \label{fig:dist}
\end{figure*}
\begin{figure*}[t!]
  \centering
  \includegraphics[width=\linewidth]{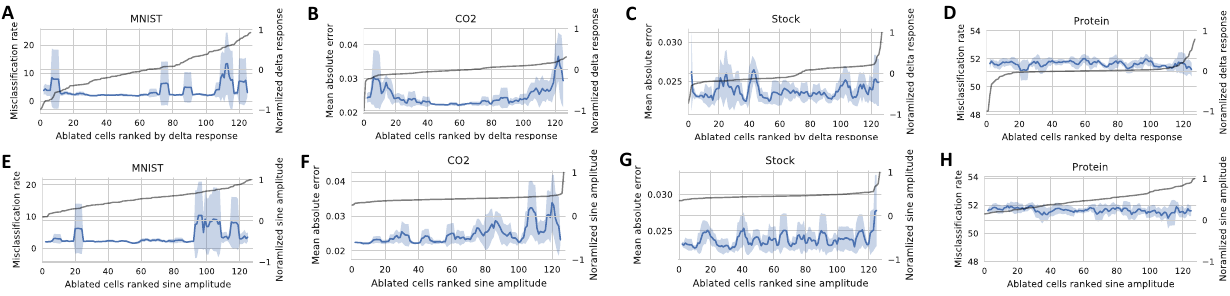}
  \caption{\textbf{Cell level ablation analysis}. Ablation of individual cells inside trained $128$ cell LSTM networks across all four datasets (left to right). Changes in the predictive error are visualized against the ranked delta response (top) and sine amplitude (bottom) of the ablated cell. The Gray solid line represents the predictions of our method (right side vertical axis) as a function of the particular response metric. The blue solid line shows the mean and the shadows represents the standard deviation of a moving average filter on the
23 ablated impact of individual neurons. This is done to highlight the trend of the ablation impact with respect to the sorted particular metric.}
  \label{fig:ablation}
\end{figure*}

Fig \ref{fig:dist}A-E, represents the distributions for the metrics sorted by the value of their specific attribute across all datasets. Cells in Protein-Net realized the fastest dynamics (i.e. smallest settling time) compared to the other networks, while realizing a similar trend to the seq-MNIST (Fig. \ref{fig:dist}A). 
The settling time distribution for the LSTM units of CO$_2$ and Stock-Net depicts cell-groups with similar speed profiles. For instance neurons 52 to 70 in Stock-Net, share the same settling time (Fig. \ref{fig:dist}A). Sine frequency stays constant for all networks except from some outliers which tend to modify their input-frequency (Fig. \ref{fig:dist}D). The delta response and the correlation metrics (Fig. \ref{fig:dist}B and Fig. \ref{fig:dist}E) both indicate the distribution of the inhibitory and excitatory behavior of individual cells within the network. Except from the Seq-MNIST net, neurons in all networks approximately keep a rate of 44\% excitatory and 56\% inhibitory dynamics. The high absolute amplitude neurons (at the two tails of Fig. \ref{fig:dist}C), are foreseen as the significant contributors to the output's decision. We validate this with an ablation analysis subsequently. 
Moreover, most neurons realize a low absolute delta-response value, for all datasets except for MNIST (Fig. \ref{fig:dist}B). This is an indication for cells with an equivalent influence on the output accuracy. Sine-amplitude stays invariant for most neurons in Stock and CO$_2$-Nets (Fig. \ref{fig:dist}C). For the seq-MNIST net and Protein-net, this distribution has a gradually increasing trend with weak values. This predicts that individual cells are globally equivalently contributing to the network's output.

\subsection{Response metrics predict significant contributing cells to the network's decision}

To assess the quality of the predictions and interpretations of the provided response characterization metrics, we performed individual cell-ablation analysis on LSTM networks and evaluated the cell-impact on the output accuracy (misclassification rate), for the classification problems and on the output performance (mean absolute error), for the regression problems. We knocked out neurons from trained LSTM networks with 128 neurons. Fig. \ref{fig:ablation}A-H illustrate the performance of the network for individual cell ablations for all four datasets. The gray solid line in each subplot, stands for the predictions of the response metrics.  

\begin{figure*}[t!]
  \centering
  \includegraphics[width=\linewidth]{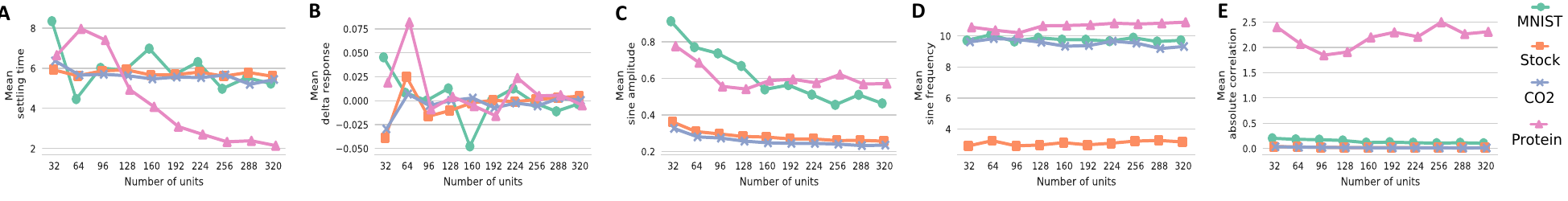}
  \caption{\textbf{Network capacity analysis.} (A-E) Response metrics as a function of the network's capacity for all datasets. The analyses illustrate how response metrics provide insights on the global network scale.}
  \label{fig:capacity}
\end{figure*}

For CO$_2$-Net, this confirms that neurons with higher sine amplitude tend to disrupt the output more (Fig \ref{fig:ablation}D). For the same network, the delta response predicted that neurons with high negative or positive value, are more significant in output's prediction. This is clearly illustrated in Fig. \ref{fig:ablation}C. For seq-MNIST-Net, the same conclusions held true; neurons with high absolute value of delta response or sine-amplitude reduce the accuracy at the output dramatically (Fig. \ref{fig:ablation}A-B). 

By analyzing the sine-amplitude and delta-response of Protein-Net, we observe that neurons are equivalently valued and tend to contribute equivalently to the output accuracy. This is verified in the ablation analysis, shown in Fig. \ref{fig:ablation}G and \ref{fig:ablation}H, where the mean-misclassification error rate stays constant for all neural ablations. The absolute value for Stock-Net was also weak in terms of these two metrics, though there were some outliers at the tails of their distribution that predicted dominant neurons. This is clearly notable when comparing the neurons 120 to 128 of Fig. \ref{fig:ablation}F to their prediction (gray line) where the amplitude of the response is maximal. In Fig. \ref{fig:ablation}E ablation experiments for neurons 1 to 40 and 100 to 128 impose higher impact on the overall output. This was also observed in the delta response prediction shown in \ref{fig:ablation}B, since neurons with stronger output response were present at the two tails of the distribution. 


\subsection{Network-level Interpretability for Trained LSTMs}


While we analyzed the response characterization distributions on a cellular level above, in this subsection we focus on the effect of network capacity on observed hidden dynamics of the system on a global scale. Reasoning on this scale allows us to draw conclusions on how increasing the expressive capacity of LSTM networks trained on the same dataset can result in vastly different learned dynamics. We experimentally vary the capacity by simply adding hidden LSTM cells to our network and retraining on the respective dataset from scratch.

The relationship between each response characteristic metric and the network capacity is visualized in Fig. ~\ref{fig:capacity}A-E. The trends across datasets are visualized in a single subplot to compare respective trends. One especially interesting result of this analysis is the capacity relationship with response amplitude (cf. Fig.~\ref{fig:capacity}C). Here we can see that the amplitude response decays roughly proportionally to $\frac{1}{N}$, for all datasets, where $N$ is the number of LSTM cells. In other words, we get the intuitive finding that as we increase the number of LSTM cells, the magnitude of each cell's relative contribution needed to make a prediction will subsequently decrease. 

Yet another key finding of this analysis is that the distribution of settling time is relatively constant across network capacity (cf. Fig.~\ref{fig:capacity}A). Intuitively, this means that the network is able to learn the underlying time delay constants represented in the dataset irrespective of the network capacity. One particularly interesting point comes for Protein-Net which exhibits vastly different behavior for both settling time (Fig.~\ref{fig:capacity}A) and correlation (Fig.~\ref{fig:capacity}E) than the remainder of the datasets. Upon closer inspection, we found that Protein-Net was heavily overfitting with increased capacity. This can be seen as an explanation for the rapid decay in its settling time as the addition of LSTM cells would increase specificity of particular cells and exhibit dynamical properties aligning with effectively memorizing pieces of the training set.

\section{Conclusion}
\label{sec:conclusion}

In this paper, we proposed a method for response characterization for LSTM networks to predict cell-contributions to the overall decision of a learned network on both the cell and network-level resolution. 
We further verified and validated our predictions by performing an ablation analysis to identify cell's which contribution heavily to the network’s output decision with our simple response characterization method. 

The resulting method establishes a novel building block for interpreting LSTM networks. 
The LSTM network's dynamic-space is broad and cannot be fully captured by fundamental input sequences. However, our methodology demonstrates that practical sub-regions of dynamics are reachable by response metrics which we use to build a systematic testbench for LSTM interpretability. We have open-sourced our algorithm to encourage other researchers to further explore dynamics of LSTM cells and interpret the kinetics of their sequential models.\footnote[1]{Code for all experiments is available online at: \url{https://github.com/raminmh/LSTM_dynamics_interpretability}} In the future, we aim to extend our approach to even more data modalities and analyze the training phase of LSTMs to interpret the learning of the converged dynamics presented in this work.

\section{Acknowledgment}
We gratefully acknowledge the support of NVIDIA Corporation with the donation of GPUs used for this research. R.M.H., M.L. and R.G. are partially supported by Horizon-2020 ECSEL Project grant No. 783163 (iDev40), and the Austrian Research Promotion Agency (FFG), Project No. 860424. A.A. is supported by the National Science Foundation (NSF) Graduate Research Fellowship Program.



\section*{Supplementary material}

# Supplementary Materials

## Observability of Recurrent Neural Networks

In this, section we recapitulate the study of the observability conditions of a recurrent neural network from a dynamic systems perspective. The study of recurrent nets and their properties as dynamic systems dates back to two decades ago where rigorous research were proposed on their weights' uniqueness (Albertini and Sontag 1994b) distinguishability (Albertini and Dai Pra 1995) and observability (Albertini and Sontag 1994a; Garzon and Botelho 1994). Here, we revisit the formal notions for recurrent networks as conditionally observable dynamic systems. A necessary condition to be able to interpret and identify a dynamic system is to check the observability property. Once a system is observable for a certain condition, then it is possible to reason about its unique internal dynamics for the realization of a particular decision region. We investigate possibilities and difficulties on exploring observability matters of LSTM networks as highly nonlinear dynamic systems. We state that finding certain regions of observability of such networks is challenging because of the highly recursive dynamics.

For defining the observability criteria, we first start with introducing some required definitions: *Independence Property* (IP) (Albertini and Sontag 1994b) – A function $\sigma$ is considered to satisfy the IP, if for any non-zero real number $w_1, ..., w_k$ and any real number $b_1, ..., b_k$, given $k \in \mathbb{Z}^+$, such that $(w_i, b_i) \neq \pm(w_j, b_j) \ \forall \ i \neq j$, the functions $1$, $\sigma(w_1 x + b_1), ..., \sigma(w_k x + b_k)$, are linearly independent. Commonly used neural network's activation functions, such as $1/(1 + e^{-x})$, $tanh(x)$ and $arctan(x)$ were formerly shown to satisfy the IP property (see more details in (Albertini and Sontag 1994b)). Note that from now on, biases $b_i$ are dropped, for brevity.

*Input Weight Restrictions* – To satisfy the observability property, the entries of the input weight matrix must be non-zero and have unique absolute values. A set of satisfactory input weight matrices, $\Omega$, can be formally presented as (Albertini and Sontag 1994a):

$$\Omega = \left\{ W \in \mathbb{R}^{n \times m} \ \middle| \ \begin{array}{ll} W_i \neq 0 & \forall \ i \in \{1..n\} \\ W_i \neq \pm W_j & \forall \ i \neq j \end{array} \right\}. \quad (1)$$

We define by $\varphi$, the set of all dynamic systems of form Eq. 1, for which $\sigma$ satisfies the IP property and $W \in \Omega$.

*Coordinate subspace* (Albertini and Dai Pra 1995) – Let's denote $c_i \ \forall_{i \in \{1..n\}}$, as the canonical basis in $\mathbb{R}^n$. A subspace $V = span\{c_{i_1}, ..., c_{i_j}\}, \ \ j > 0$, determines the coordinate subspace of the system. Coordinate subspaces are invariant to all projections and their sum will also get realized in the same form. For a dynamic neural network described in the form of Eq. 1, for every pair (W,C), there exist a unique largest R-invariant coordinate subspace, $\vartheta_c(R, C)$, included in the kernel of C (Albertini and Sontag 1994a). We now have all the definitions required to investigate observability.

## Observability of Vanilla Recurrent Neural Networks

A dynamic system is observable if there is some input sequence that gives rise to distinct outputs for two different initial states at which the system is started (Sontag 2013). Observability is a necessary condition for the interpretation of a system's kinetics, and controllability (Albertini and Dai Pra 1995). Therefore, if we can satisfy the observability property, one can analytically reason about the unique weight spaces (see (Albertini and Sontag 1994b)) that realize certain decision regions. Given the properties and definitions in the previous section, observability of a recurrent neural network of the form of Eq. 1, can be determined using the following theorem:

**Theorem 1** *For a dynamic system $\xi$ as a subset of $\varphi$, if and only if $\vartheta_c(R, C) = 0$ (this is equivalent to $\ker R \cap \ker C = 0$, then $\xi$ is observable (see the proof in (Albertini and Sontag 1994a; Sontag 2013)).*

$\vartheta_c(R, C) = 0$ interprets as there is no R-invariant coordinate subspace which is included in kernel of C. It is shown in (Sontag 1979), that if all the columns of C is nonzero, then $\vartheta_c = 0$. For the commonly used RNNs $C$ is usually determined as the identity matrix, where the states of the hidden layers, $h$ are linearly projected to the output. If $R$ is invertible (det $R \neq 0$), then $\ker R \cap \ker C = 0$.

Therefore one can reform the observability condition for the hidden states of a recurrent net, $\xi$, as:

*Let $\xi \in \varphi$, if det $R \neq 0$ and every column of $C$ is nonzero, then $\xi$ is considered observable.*

**Observability of LSTM Networks** All gates realize a feed-forward dynamics as shown in Eq. 6, with an $R = 0$ for a dynamic system described in Eq. 1. The gating dynamics, stand-alone, are therefore denoted as: $\dot{h} = \sigma(Wx)$, where $\dot{h}$ represents the current decision of the gating behavior. This system is observable, since a projection map $f : \mathbb{R}^n \to \mathbb{R}^n$ with bounds, is observable if and only if the function realizes continuity at every point within its range (Garzon and Botelho 1994). This is true for the input block of the cell in isolation as well, since $\tanh$ is also a continuous function it its range. By shutting down all the gates, the hidden state of the LSTM network follows similar dynamics as in standard RNNs. With the difference of having the state passing through two $\tanh$ nonlinearities as $\dot{y} = \tanh(\tanh(R_z y + W_z x))$. Since $\tanh(\tanh(x))$ is a real analytic function, and is bound, same principles and also Theorem 1 are applicable regarding the network's observability property. The cell-state $c$ and the hidden-state dynamics of the LSTM, $y$ form the actual dynamics of the system. $c$ is composed of a multiplicative nonlinearity expressed by two observable systems $i$ and $z$. The outcome of this system is a bounded function in $\mathbb{R}^n$ and is analytic for all coordinate subspaces. If a given set of $R$ and $W$ matrices, satisfy the conditions of the Theorem 1, this term of the equation is observable. The forget-gate, $f$, introduces additional nonlinearity to the cell-state kinetics. The resulting function (second term in Eq. 4), is not bound although analytic. Therefore, there are weight spaces which makes the observable regime for the system narrower. The hidden-state dynamics, $y$, standalone, similar to the first term in Eq. 4, realizes observable dynamics. However, the observability of the cell is limited to the cell-state observable regions of dynamics which are challenging to define analytically, due to highly nonlinear partial dependencies. On this basis, we introduced a quantitative empirical method to find out sub-regions of dynamics by using response metrics in order to interpret LSTM's dynamics.

## Supplementary Figures

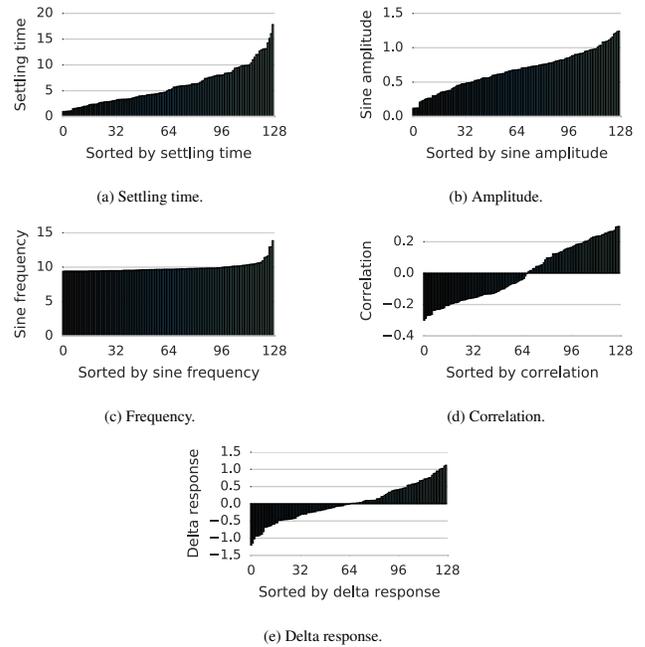

Figure S1: Seq-MNIST- Cell response characteristics

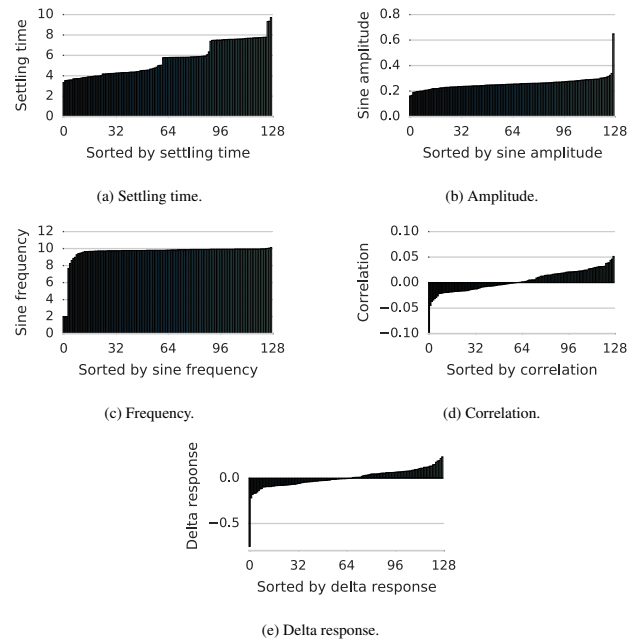

Figure S2: $CO_2$- Cell response characteristics

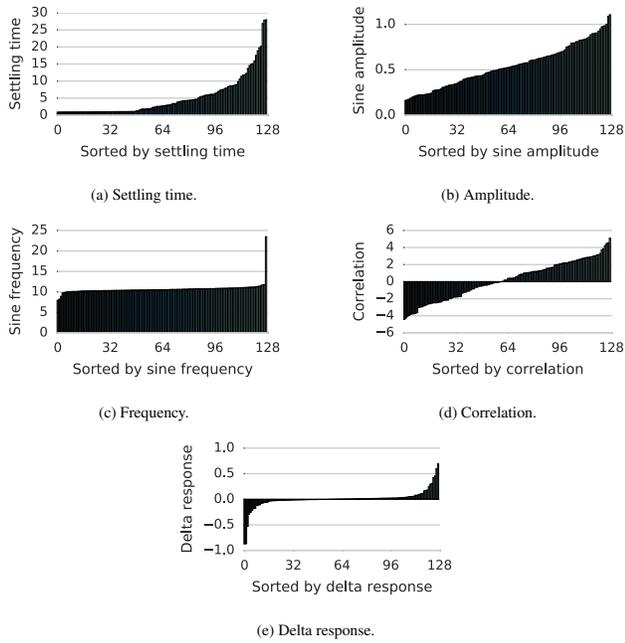

Figure S3: Protein - Cell response characteristics

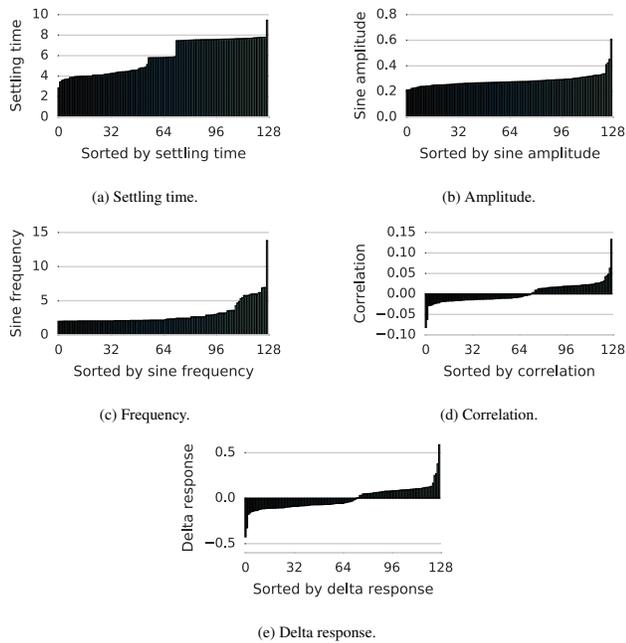

Figure S4: Stock - Cell response characteristics

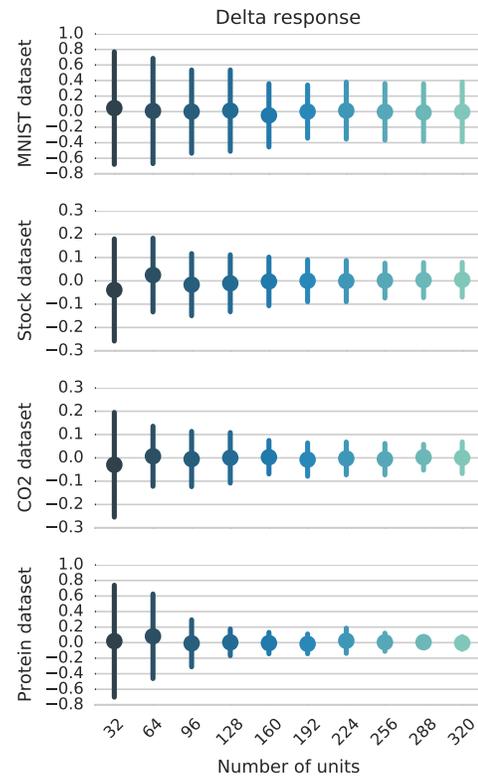

Figure S5: Delta response over capacity. We can observe that for all datasets the standard deviation of the delta response decreases with higher capacities of the network. This makes sense since each neuron can become more specific in terms of how much it contributes to the overall output.

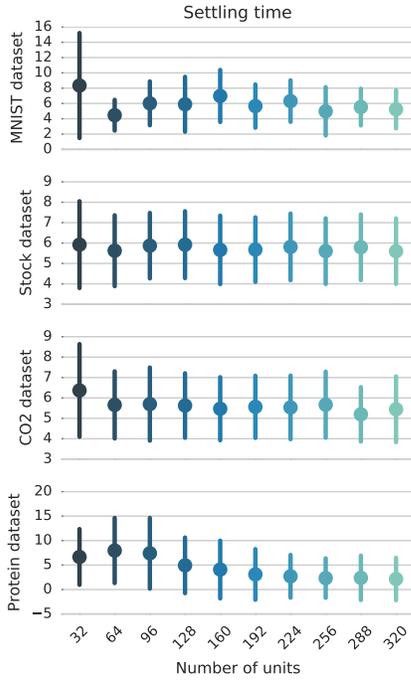

Figure S6: Settling time over capacity.

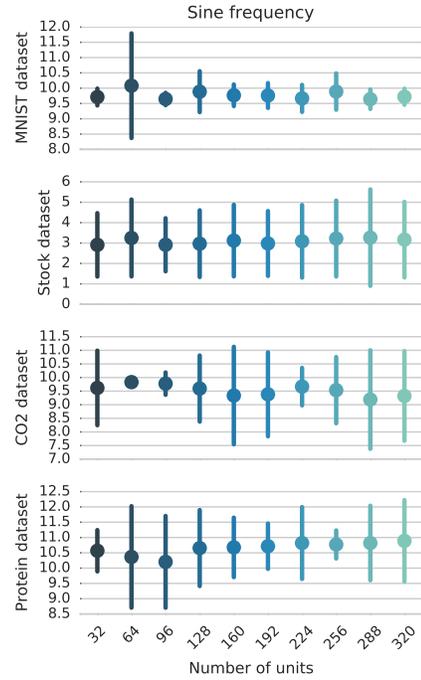

Figure S8: Sine-Frequency over capacity.

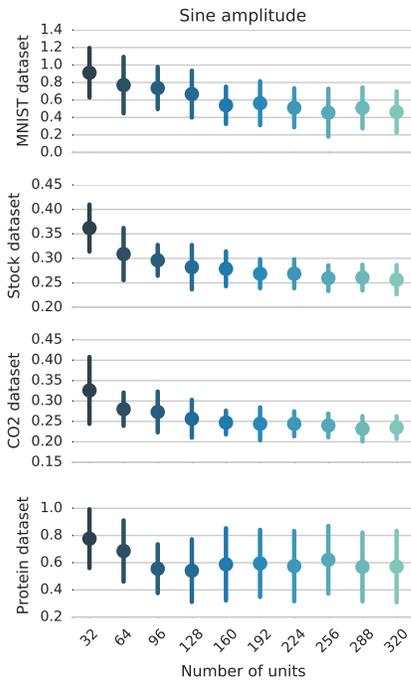

Figure S7: Sine-Amplitude over capacity.

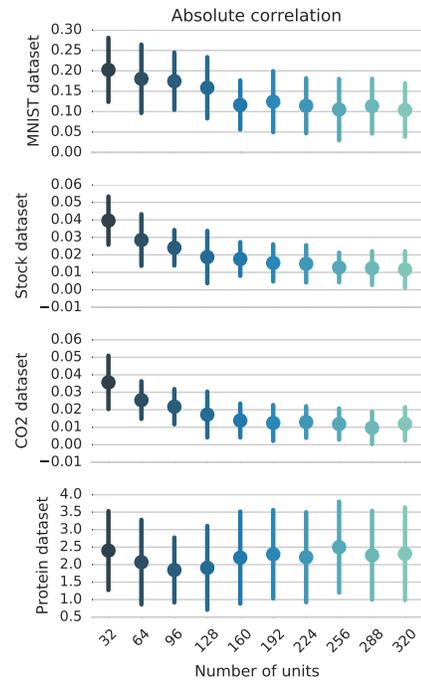

Figure S9: Correlation over capacity.



\end{document}